\documentclass[runningheads]{llncs}
\usepackage{graphicx}
\usepackage{amsmath}
\usepackage{hyperref}
\usepackage{color}

\newcommand{\arrow}[1]
        {\, {\auxarrow\limits^{#1}} \,}
\newcommand{\auxarrow}
	{\mathop{- \!\!\!\! \longrightarrow}}
\newcommand{\infr}[2]
	{\renewcommand{\arraystretch}{1.5}
	\begin{array}{c}
	#1\\
	\hline
	#2
	\end{array}}

\begin{document}
\title{A process algebraic framework for multi-agent dynamic epistemic systems}
%
%
\author{Alessandro Aldini\orcidID{0000-0002-7250-5011}}

\institute{
University of Urbino Carlo Bo, Italy \\
\email{alessandro.aldini@uniurb.it}
}
\maketitle              
\begin{abstract}
This paper combines the classical model of labeled transition systems with the epistemic model for reasoning about knowledge. The result is a 
unifying framework for modeling and analyzing multi-agent, knowledge-based, dynamic systems. On the modeling side, we propose a process algebraic, agent-oriented specification language that makes such a framework easy to use for practical purposes. On the verification side, we define a modal logic encompassing temporal and epistemic operators.
\end{abstract}

\section{Introduction}

The formal modeling of agent-based systems and the knowledge transfer enabled by the related interactions is a research field common to several areas, ranging from concurrency theory to epistemic logic. 

In the former setting, two basic models are mainly adopted to describe the dynamics of systems: 
$(i)$ \textit{Kripke structures} are graphs where the nodes are annotated with atomic propositions stating what is true in the system state associated with the node, and $(ii)$ \textit{labeled transition systems} (LTSs) are graphs where the arcs are annotated with actions representing the events causing a change of system state. 
Both paradigms are equipped with temporal logics for the description of properties, like, e.g., Computation Tree Logic (CTL) for state-based structures~\cite{10.5555/1373322} and Hennessy-Milner Logic (HML) for action-based systems~\cite{HML}. 

In the latter setting, the focus is on reasoning about knowledge from the viewpoint of non-omniscient agents in terms of their capability of distinguishing different scenarios \cite{DHHK2015}. The standard way to model epistemic notions is through a state-based epistemic model called \textit{Kripke model}. 
Every state (called \textit{possible world}) is characterized by the propositional statements that hold in it, as in Kripke structures. At the same time, an \textit{accessibility} relation determines, from the viewpoint of the agent under consideration, which worlds are compatible (\textit{indistinguishable}) with her knowledge in the current world. In this setting,
epistemic logic introduces a knowledge modality for 
reasoning about what agents know or can deduce from the information at their disposal and, possibly, for tracking the information flow among agents. 

The connections between the two strands of research are evident and, in some cases, the mutual foundational influence between traditional concurrency models and epistemic models is investigated (see, e.g., \cite{10.5555/2340820.2340835,10.1007/978-3-642-32940-1_23,GUZMAN2017107,10.1007/978-3-540-75560-9_18}). Specific examples of cross-fertilization can be found in the formal analysis of security protocols; see \cite{Dechesne2007} for a survey and, in particular,
\cite{10.1007/978-3-642-02138-1_12,10.1145/2166956.2166962,10.1007/978-3-030-50086-3_7,BAVENDIEK2022100735}, where logical formalizations of knowledge are integrated into a modeling framework based on pi-calculus in order to characterize the intruder's capability of breaking security properties.

The main goal of this paper is to combine the advantages of the two approaches by merging in a novel, multi-agent framework the capability of the LTS-based semantics of modeling dynamic, temporal behaviors with the capability of the epistemic models of representing what agents know or do not know. 
The rough idea behind the combination is associating a Kripke model with each state of an LTS.
As additional contributions, this novel framework is enriched with a logic including dynamic and epistemic modalities and a high-level, process-algebraic specification language.

In the following, we introduce the model of Kripke labeled transition systems (Section \ref{sect:klts}) as a combination of epistemic models and LTSs. We define a logic for describing properties for such a model and establish the equivalence relation that is characterized by the logic.
Then, we propose a process algebraic language for modeling agent-oriented concurrent systems with semantics based on Kripke labeled transition systems (Section \ref{sect:language}). To emphasize the usability of this language, we describe a case study based on a popular, classical board game (Section \ref{sect:cluedo}). 
Finally, we discuss related work and potential future directions
(Section \ref{sect:rw}).

\section{Kripke labeled transition systems}
\label{sect:klts}

Let $\mathcal{A}$ be a set of agents (ranging over $i,j,\ldots$), $Act$ a set of actions (ranging over $\pi,\pi',\ldots$), and $At$ a set of atomic propositions (ranging over $p,q,\ldots$); we will use $X, Y, \ldots$ to denote subsets of $At$. First of all, we recall the definitions of labeled transition system and multi-agent epistemic model.

\begin{definition}
A labeled transition system (LTS) is a tuple $(S,T,s_0)$ where $S$ is a non-empty set of states (with $s_0$ the initial state) and $T \subseteq S \times Act \times S$ is the action-labeled transition relation.
\end{definition}

In the setting of computation modeling,
LTSs describe the evolving behavior of discrete systems, where the actions labeling the transitions represent events leading from one configuration of the system to another.

\begin{definition}
A multi-agent epistemic model (called Kripke model) is a tuple $(S,\{R_i \mid i \in \mathcal{A} \},v)$, where $S$ is a non-empty set of states; for every $i \in \mathcal{A}$, $R_i \in 2^{S \times S}$ is a binary (accessibility) relation over $S$; $v: S \to 2^{At}$ is a valuation function assigning to each state the set of propositions that hold in the state.
\end{definition}

A pointed (resp., rooted) Kripke model is a pair $((S,\{R_i \mid i \in \mathcal{A} \},v), s)$, where $s \in S$ is the current (resp., initial) state. 
Kripke models serve as the basis of the semantics for various modal logics and, in the case of epistemic languages, allow us to reason about knowledge in terms of information accessibility.

For our purposes, combining the dynamic action-based nature of LTSs with the possible worlds description of Kripke models results in action-based systems, the states of which are associated with accessibility relations and valuations. 

\begin{definition}
A Kripke labeled transition system (KLTS) is a tuple $M = (S, T, \linebreak \{r_i \mid i \in \mathcal{A}\}, v )$, where
$S$ is a non-empty set of states; $T \subseteq S \times Act \times S$ is a transition relation; for every $i \in \mathcal{A}$, $r_i : S \rightarrow 2^{2^{At} \times 2^{At}}$ is a function mapping each state to a binary (accessibility) relation over $2^{At}$; $v: S \to 2^{At} $ is a valuation function.
\end{definition}

Firstly, states should not be considered dependent on atomic propositions. They are primitive semantic objects so that the set of propositions satisfied by a state does not uniquely identify the state. Secondly, each accessibility relation $r_i(s)$ relates elements of $2^{At}$ and expresses the actual observational power of agent $i$ in state $s$ with respect to the truth of the propositions in $At$. In other words, $r_i(s)$ describes the distinguishing power of agent $i$ in $s$, intended as her capability of distinguishing the possible worlds identified by the values of the propositions. 
Under the indistinguishability interpretation of epistemic logic, $r_i(s)$ expresses informational indistinguishability between possible worlds. 
More precisely, $(X,Y) \in r_i(s)$ means that in $s$ the agent $i$ has insufficient information to establish whether we are in a state in which all and only the propositions of $X$ hold or in a state in which all and only the propositions of $Y$ hold. 
Hence, both $X$ and $Y$ are compatible with the knowledge of the agent $i$ in $s$. By virtue of this interpretation, in the following we assume that the accessibility relations are equivalence relations. Thirdly, the transition relation $T$ and the valuation function $v$ are interpreted as usual.

\begin{example}
If $(\{p\} \cup X, \{p\} \cup Y)$ belongs to $r_i(s)$ for any choice of $X,Y \in 2^{At}$, then, in $s$,
all the possible worlds in which $p$ holds are mutually indistinguishable from the viewpoint of agent $i$. If we also have that $(\{p\} \cup X,Y) \not\in r_i(s)$ whenever  $p \not\in Y$, we conclude that agent $i$ distinguishes all and only the pairs of worlds
differing for the valuation of $p$. Later, we will realize that this means that, in $s$, agent $i$ \textit{knows} the truth value of $p$ and is ignorant of any other proposition.
\end{example}

%
 
\begin{remark}
From a KLTS, an LTS can be derived. In particular,
if we omit from a rooted KLTS $((S,T,\_, \_),s_0)$ the accessibility relations and the valuation function, we obtain an LTS. Moreover, the KLTS $(2^{At},\emptyset,\{r_i \mid i \in \mathcal{A} \}, \mathit{id})$ -- where \textit{id} is the identity function and $r_i(s) = r_i(s')$ for all $i \in \mathcal{A}$ and for any $s,s' \in 2^{At}$ -- is a Kripke model. 
\end{remark}

LTSs and Kripke models provide the semantics for interpreting properties expressed in various modal logics. Inspired by temporal logics and epistemic logics, we propose a modal logic that naturally combines temporal and epistemic ingredients, called Kripke Temporal (KT) logic.

\begin{definition}[KT Logic]
The language $\mathcal{L}_{KT}$ of the KT logic is defined by the following two-layers grammar:
\[\begin{array}{lcl}
\phi & \rightarrow & \top \mid p \mid \neg \phi \mid \phi \land \phi 
\mid \langle \pi \rangle \phi \mid \psi \\
\psi & \rightarrow & \top \mid p \mid \neg\psi \mid \psi\land\psi \mid K_i \psi
\end{array}\]
\end{definition}

The $\psi$ formulas are called epistemic formulas.
Note that the KT logic results from the combination and encompasses both HML \cite{HML} and Epistemic Logic \cite{DHHK2015}. 

\begin{definition}
Given a KLTS $M = \{ S, T, \{ r_i \mid i \in \mathcal{A}\}, v \}$ and denoted $M_s = (2^{At},\{r_i(s) \mid i \in \mathcal{A}\},id)$, with $s \in S$, the truth of $\varphi \in \mathcal{L}_{KT}$ at $s \in S$, denoted $M,s \models \varphi$, is defined as follows:
\[
\begin{array}{ll}
1.~M,s \models \top & 2.~M,s \models p ~\mathit{iff}~ p \in v(s) \\
3.~M,s \models \neg \varphi ~\mathit{iff}~ M,s \not\models \varphi \hspace{2mm} &
4.~M,s \models \varphi_1 \land \varphi_2 ~\mathit{iff}~ M,s \models \varphi_1 
~\mathit{and}~ M,s \models \varphi_2 \\
\multicolumn{2}{l}{5.~M,s \models \langle \pi \rangle \phi ~\mathit{iff}~ \exists s' \ldotp (s,\pi,s') \in T ~\mathit{and}~ M,s' \models \phi} \\
\multicolumn{2}{l}{6.~M,s \models K_i \psi ~\mathit{iff}~
M_s, v(s) \models_\mathrm{K} K_i \psi,~\textit{where the relation } \models_\mathrm{K} \textit{ is defined as:} } \\
\multicolumn{2}{l}{ \qquad \qquad
\begin{array}{l}
(a)~M_s,X \models_\mathrm{K} \top \\ 
(b)~M_s,X \models_\mathrm{K} p~\mathit{iff}~p \in X \\
(c)~M_s,X \models_\mathrm{K} \neg \psi~\mathit{iff}~ M_s,X \not\models_\mathrm{K} \psi \\
(d)~M_s,X \models_\mathrm{K} \psi_1 \land \psi_2 ~\mathit{iff}~ M_s,X \models_\mathrm{K} \psi_1 ~\mathit{and}~ M_s,X \models_\mathrm{K} \psi_2 \\
(e)~M_s,X \models_\mathrm{K} K_i \psi ~\mathit{iff}~ \forall Y \ldotp (X,Y) \in r_i(s): M_s,Y \models_\mathrm{K} \psi
\end{array}}
\end{array}
\]
\end{definition}
Note that the formula $K_i \psi$ is evaluated in $s \in S$ with respect to the accessibility relations associated with $s$, thus emphasizing the view of the KLTS as an LTS with a Kripke model $M_s = (2^{At},\{r_i(s) \mid i \in \mathcal{A}\},id)$ for each state $s$. Hence, the semantics of an epistemic formula evaluated in $s$ depends on such a Kripke model ($\models_\mathrm{K}$ is the classical satisfiability relation for Kripke models).
By virtue of the indistinguishability interpretation we adopted, since we are assuming to work with accessibility relations that are reflexive, symmetric, and transitive, the reference system for the knowledge modality is S5~\cite{blackburn2002modal}. 

Based on the semantics above, two states $s$ and $s'$ are modal equivalent, denoted $s \equiv s'$, if and only if they satisfy the same formulas. 
The KT logic characterizes the following notion of behavioral equivalence.


\begin{definition}\label{def-bisimulation}
Let $M = (S,T,\{r_i \mid i \in \mathcal{A}\},v)$ be a KLTS.
A binary equivalence relation $B$ on $S$ is a bisimulation iff whenever $(s,t) \in B$ then:
\begin{enumerate}
    \item $v(s) = v(t)$;
    \item if $(s,a,s') \in T$ then $\exists t' \ldotp (t,a,t') \in T$ and $(s',t') \in B$;
\item there exists a binary equivalence relation $\mathcal{B}_{st}$ between the worlds of the Kripke models $(2^{At},\{r_i(s) \mid i \in \mathcal{A}\},\mathit{id})$ pointed at $v(s)$ and $(2^{At},\{r_i(t) \mid i \in \mathcal{A}\},id)$ pointed at $ v(t)$, such that $(v(s),v(t)) \in \mathcal{B}_{st}$ and for any $X,Y \in 2^{At}$, whenever $(X,Y) \in \mathcal{B}_{st}$ then:
\begin{itemize}
    \item $X = Y$;
    \item if $(X,X') \in r_i(s)$ for $i \in \mathcal{A}$, then $\exists Y' \ldotp (Y,Y') \in r_i(t)$ and $(X',Y') \in \mathcal{B}_{st}$.
\end{itemize}
\end{enumerate}
\end{definition}

Note that conditions $1.$ and $3.$ resemble the definition of modal bisimulation for Kripke models~\cite{blackburn2002modal}, while condition $2.$ characterizes the strong bisimulation for LTSs~\cite{HML}.
Two states $s$ and $s'$ are bisimilar, denoted 
$s \sim s'$, if and only if there exists a bisimulation $B$ such that $(s,s') \in B$.
The correspondence theorem relates bisimilar states and equivalent states whenever the KLTS is image finite, i.e., for all states and actions, the image of $s$ (under any accessibility relation) and the image of $s, \pi$ (under the transition relation) are finite.

\begin{theorem}\label{main_thm}
For any image-finite KLTS, $\sim$ coincides with $\equiv$.
\end{theorem}

As a consequence of the grammar structure and the semantics of the KT logic, decidability and verification algorithms are inherited from the results related to epistemic logic and HML.


\section{A language for Kripke labeled transition systems}
\label{sect:language}

In this section, we define a process-algebraic, agent-oriented language with value passing, the semantics of which is given in terms of KLTSs.
We start by defining a basic calculus (see, e.g., \cite{Fokkink2007,GorrieriVersaribook}) with value passing (see, e.g., \cite{Hennessy1991,Huang2012ValuepassingCW}) for the description of sequential process terms. 
Let $A$ be a set of action names (ranging over $a,b,\ldots$) including the special action names $\tau$ and \textit{set}. To model value passing, we will use variables ($x,y, \ldots,f,g,\ldots$), values ($v, v', \ldots$) from fixed domains, and expressions ($e, e', \ldots$) that usually represent simple values.

\begin{definition}
The set $\mathcal{L}$ of process terms of the  calculus for sequential processes is generated through the following syntax:
\[\begin{array}{l}
P \rightarrow \underline0 \mid \sum_{k \in I} \pi_k \, . \, P_k \mid C(e_1, \ldots, e_n)  \\
\pi \rightarrow b \mid a(y,f) \mid \Bar{a}(i,\psi) \mid \mathit{set}(p,w) \\ 
\end{array}\] 
where $b \in A \backslash \{\mathit{set}\}$, $a \in A \backslash \{\tau,\mathit{set}\}$, $I$ is any finite indexing set, $w$ is a boolean value, $C$ is a constant name with the natural $n \ge 0$ being the arity of $C$.
\end{definition}
The constant $\underline0$ stands for the inactive, halted process. 
The summation operator represents a nondeterministic choice enacting one of the guarded process terms $\pi_k . P_k$, which executes action $\pi_k$ and then behaves as process term $P_k$ (we will use $E$ to denote a non-empty summation). The constant $C$ is used to express recursive processes with $n \ge 0$ parameters, and must be associated with a defining equation of the form $C(x_1, \ldots, x_n) := P$. 
The notation $\pi$ stands for any action, which can be an internal action $b$, an input action $a(y,f)$, an output action $\Bar{a}(i,\psi)$, or an assignment action $\mathit{set}(p,w)$. 

An assignment action has the effect of setting the proposition $p$ to the boolean value $w$.
An output action communicates an epistemic formula $\psi$ to the agent $i$, while an input action receives a formula assigned to the variable $f$ from an agent assigned to the variable $y$.

As usual in calculi with value passing, each occurrence of any variable in a process term $P$ is bound by either an input action or a constant definition.
For instance, $x$ is bound in $C(x) := \Bar{a}(x,p \land q) \,.\, C(x+1)$
and in $a(x,f) \,.\, \Bar{b}(x,\top) \,.\, \underline0$, but not in
$\Bar{a}(x,p \land q) \,.\, \underline0$.
Moreover, we write $i/x$ and $\psi/f$ for substitutions of values for variables, and 
denote by $P[i/x,\psi/f]$ the result of substituting $i$ (resp., $\psi$) for all free (not bound) occurrences of $x$ (resp., $f$) in $P$.  

Formally, the behavior of a process term $Q$ is described in structural operational semantics style as the LTS rooted at $Q$ and defined by the transition relation $T \subseteq \mathcal{L} \times \mathit{Act} \times \mathcal{L}$ that is the least transition relation generated by the axioms and the rules in Table~\ref{tab:sem_seq}. 
All the pre- and post-conditions associated with knowledge-based behaviors (i.e., communications and assignments) will be defined when introducing the parallel composition of process terms and the knowledge structures.

\begin{table}[tb]
\caption{Semantics rules for sequential processes \label{tab:sem_seq}}
\[\begin{array}{|c|}
\hline 
\mathit{(prefix)} \quad \quad b \, . \, P \arrow{b} P 
\quad \quad
\mathit{set}(p,w) \, . \, P \arrow{\mathit{set}(p,w)} P
\quad \quad
\Bar{a}(i,\psi) \, . \, P \arrow{\Bar{a}(i,\psi) } P
\\
\mathit{(input)} \quad \quad 
a(y,f) \, . \, P \arrow{a(i,\psi)} P[i/y,\psi/f] \quad \mathit{\ for\ any} \ i \in \mathcal{A} \mathit{\ and\ epistemic\ formula\ } \psi 
\\
\mathit{(sum)} \quad \quad \infr{\pi \, . \, P \arrow{\pi} P}{ \pi \,.\, P + E
\arrow{\pi} P} \\
\mathit{(recursion)} \quad \quad
\infr{P[v_1/x_1,\ldots,v_n/x_n] \arrow{\pi}{} P'}{C(e_1,\ldots,e_n) \arrow{\pi}{} P'} \quad 
\begin{array}{l}
C(x_1,\ldots,x_n) := P \ \mathrm{and} \\[-2mm]
\mathrm{each}\ e_i\ \mathrm{evaluates\ to\ } v_i
\end{array}
\\ \hline
\end{array}\]
\end{table}

\begin{example}
The process term
$\mathit{Agent} := 
\mathit{receive}(y,f) . \overline{\mathit{send}}(y+1,f) .
\mathit{Agent}$
represents an agent without parameters that is available to receive as input a formula from an agent $y$, and then forwards such a formula as an output to agent $y+1$ (here, we assume that identities are naturals).
\end{example}

\subsection{Agents and pool of agents}

Process terms represent behavioral patterns of agents, while an agent is an instance of a process term with a unique identity. Several agents may communicate with each other to form a network of agents. Hence, we need to formalize the notion of agent and how agents interact in a so-called pool of agents. A dynamic knowledge structure will be added to regulate such interactions.

Agents are described by tuples of the form $\langle i \in \mathcal{A}, P \in \mathcal{L} \rangle$ and are ranged over by $\mathcal{I},\mathcal{J},\ldots$.
The semantics of $\langle i, P \rangle$
is given by the LTS expressing the behavior of $P$, up to the renaming of the actions as defined by the semantic rule:
\[\begin{array}{c}
\mathit{(agent)} \quad \quad \infr{P \arrow{\pi}{} P'}{\langle i,P \rangle \arrow{i.\pi}{} \langle i,P' \rangle}
\end{array}\]

So far, 
we abstracted from the interaction among agents and the underlying knowledge base. Now, we combine the behavior of several agents by integrating the notion of knowledge, which will allow us to specify how they can interact.

\begin{definition}
A pool of agents is a tuple $\mathcal{P} = (\cup_i \mathcal{I}_i, R, X)$ where: 
\begin{itemize}
    \item $\cup_i \mathcal{I}_i$ is a finite set of agents;
    \item $ R = \{ R_i \mid i \in \mathcal{A} \}$ where each $R_i$ is a binary accessibility relation over $2^{At}$; 
    \item $X \subseteq{At}$ is the set of true propositions.
\end{itemize}
\end{definition}

The behavior of the set $\cup_i \mathcal{I}_i$ depends on the behavior of each $\mathcal{I}_i$ and is defined as an element of the cartesian product $(\mathcal{A}\times\mathcal{L})^n$, where $n$ denotes the cardinality (i.e., the number of agents) of the pool.
Then, each accessibility relation $R_i$ expresses the capability of the agent $i$ to distinguish the possible worlds based on the values that can be attributed to the propositions of $\mathit{At}$. Finally, set $X$ denotes the current truth assignment for the propositions of $\mathit{At}$.

The agents described by a tuple $\mathcal{P}$ can perform actions, either synchronously or autonomously, thus making the system dynamic. On the one hand, the internal actions that are not related to knowledge in any way and the assignment actions represent the autonomous actions of agents. On the other hand, input and output actions represent synchronous actions that express knowledge transfer between agents.

Formally, such a joint knowledge-based and action-based behavior is represented by a KLTS describing the evolution of the pool of agents.

\begin{definition}\label{def-pool}
Let $\mathcal{P} = (\cup_i \mathcal{I}_i, \{ R_i \mid i \in \mathcal{A} \}, X)$ be a pool of agents of cardinality $n$. The semantics of $\mathcal{P}$ is given by the KLTS $((S,T,\{r_i \mid i \in \mathcal{A}\},v), \mathcal{P})$
rooted at $\mathcal{P}$, where $S$ is the set of pool tuples, $T$ is the least transition relation generated by the rules of Table~\ref{tab:sem_pool}, and for each state $s = (\_,\{ R^s_{i} \mid i \in \mathcal{A} \},X_s) \in S$ it holds that $r_i(s) = R^s_{i}$ for each $i \in \mathcal{A}$ and $v(s) = X_s$.
\end{definition}

\begin{table}[tb]
\caption{Semantics rule for a pool of agents \label{tab:sem_pool}}
\[\begin{array}{|c|}
\hline
\mathit{(pool)}  \quad \quad \infr{\mathcal{J} \arrow{j.b}{} \mathcal{J}'}{ (\cup_i \mathcal{I}_i \cup\mathcal{J}, R, X) \arrow{j.b}{} 
(\cup_i \mathcal{I}_i \cup\mathcal{J'}, R, X)} 
\\[6mm]
\mathit{(set)} \quad \quad \infr{\mathcal{J} \arrow{j.\mathit{set(p,w)}}{} \mathcal{J}'}{ (\cup_i \mathcal{I}_i \cup\mathcal{J}, \cup_i R_i \cup R_j, X) \arrow{\tau}{} 
(\cup_i \mathcal{I}_i \cup\mathcal{J'}, \cup_i R'_i \cup R'_j, X')} 
\\[6mm] 
\begin{array}{l}
\mathrm{where}\ X' = \left \{
\begin{array}{ll}
X \backslash \{p\} & \mathrm{if}\ w=0 \\
X \cup \{p\} & \mathrm{if}\ w=1 \\ 
\end{array} 
\right. \
\mathrm{and,\ for}\ N = (2^{At},  \cup_i R_i \cup R_j, id): \\[3mm]
-\ R_j' = R_j \backslash \{ (Y,Y') \mid \mathit{diff}(N,Y,Y',p) \} \\[1mm]
-\ R_i' = \mathit{closure}(R_i \cup \{ (\{p\} \cup Y, Y) \mid p \not\in Y \} \cup \{ (Y, \{p\} \cup Y) \mid p \not\in Y \} )
\end{array}
\\[8mm]
\mathit{(com)} \quad \quad  \infr{(\cup_k \mathcal{I}_k \cup\mathcal{I} \cup\mathcal{J}, R \cup R_j, X) \models K_i \psi \quad \mathcal{I} \arrow{i.\mathit{\Bar{a}(j,\psi)}}{} \mathcal{I}' \quad \mathcal{J} \arrow{j.\mathit{a(i,\psi)} }{} \mathcal{J}'}{ (\cup_k \mathcal{I}_k \cup\mathcal{I} \cup\mathcal{J}, R \cup R_j, X) \arrow{\tau}{} 
(\cup_k \mathcal{I}_k \cup\mathcal{I'} \cup\mathcal{J'}, R \cup R_j', X)} \\[6mm]
\mathrm{where,\ for\ } N = (2^{At}, R \cup R_j, id):
R_j' = R_j \backslash \{ (Y,Y') \mid \mathit{diff}(N,Y,Y',\psi) \}
\\[2mm] \hline \\[-2mm]
\begin{array}{l}
\mathit{diff}(N,X,Y,\psi) = (N,X \models_{\mathrm{K}} \psi \land N,Y \not\models_{\mathrm{K}} \psi) \lor 
(N,X \not\models _{\mathrm{K}}\psi \land N,Y \models_{\mathrm{K}} \psi) \\
\mathit{closure}(R_j) = R_j \cup \{ (X,Y) \mid \exists Z \ldotp (X,Z) \in R_j \land (Z,Y) \in R_j \} 
\end{array}
\\ \hline
\end{array}\]
\end{table}

We now illustrate the rules of Table~\ref{tab:sem_pool}.
The rule ($\mathit{pool}$) describes the asynchronous execution of autonomous actions of the form $b \in A \backslash\{ \mathit{set} \}$ by any agent of the pool. Note that such actions do not change the knowledge structure modeled by the accessibility relations and the truth assignment.

The rule ($\mathit{set}$) describes the asynchronous execution of autonomous actions of the form $\mathit{set}(p,w)$ by any agent $j$, whose side effect is that the truth assignment $X$ associated with the current tuple is updated according to the assignment $p = w$ (see the definition of $X'$). The accessibility relations are also updated. On the one hand, the agent $j$ performing the assignment acquires knowledge (if not yet possessed) of $p$. Hence, in $R_j$, all the possible worlds differing for the valuation of $p$ (see function \textit{diff}) cannot be mutually accessible anymore, as they are distinguishable by the value of $p$. 
Note that, as we will show, such suppression of connections ensures that the accessibility relation remains an equivalence.
On the other hand, all the other agents $i \not= j$ lose knowledge (if previously possessed) of $p$, as the assignment is not considered public (as emphasized by the fact that the resulting action is a silent action $\tau$). Therefore, in each accessibility relation of those agents, all the possible worlds differing only for the valuation of $p$ must become mutually accessible, as they cannot be distinguished anymore. 
Note that such addition of connections considers the symmetric pairs and, through the \textit{closure} operation, the transitive relations, thus ensuring, as we will show, that the accessibility relation remains an equivalence.

The most interesting rule is ($\mathit{com}$), which expresses a communication from an output to a corresponding input (the two actions refer to the same action name $a$). The agent $i$ performing the output and the agent $j$ performing the corresponding input synchronize, i.e., they both advance simultaneously. However, the resulting synchronization is enabled only if the epistemic formula $\psi$ communicated from $i$ to $j$ is known by $i$. If this is the case, $j$ acquires knowledge of $\psi$, and the accessibility relation $R_j$ is updated accordingly. In fact, agent $j$ becomes able to distinguish those possible worlds that differ from each other for the evaluation of $\psi$.
The communication is private (the synchronization result is a silent action $\tau$), i.e., the knowledge transfer involves only the agent $j$ and no one else.

\begin{lemma}
The KLTS modeling the behavior of a pool of agents is image finite.
\end{lemma}

This result immediately derives by Definition~\ref{def-pool} and the semantics of Table~\ref{tab:sem_pool}.

As anticipated, an important result to show is that the semantics of Table~\ref{tab:sem_pool} preserves the indistinguishability interpretation of the accessibility relations.

\begin{theorem}\label{rel-thm}
Let $\mathcal{P} = (\cup_i \mathcal{I}_i, \{ R_i \mid i \in \mathcal{A} \}, X)$ be a pool of agents such that each $R_i$, with $i \in \mathcal{A}$, is a $\mathtt{P}$-relation (for $\mathtt{P}$ in \{\textit{reflexive}, \textit{symmetric}, \textit{transitive}\}) and $((S,T,\{r_i \mid i \in \mathcal{A}\},v), \mathcal{P})$ be the semantics of
$\mathcal{P}$. Then, for each $i \in \mathcal{A}$ and for each $s \in S$, it holds that $r_i(s)$ is a $\mathtt{P}$-relation.

\end{theorem}


\section{Use case: playing Cluedo}
\label{sect:cluedo}

The present case study is designed to highlight the modeling features and analysis opportunities of our framework. Despite its simplicity, this use case encompasses many of the features of real-world applications, including strategic thinking, private and public communications, and knowledge transfer.
For the sake of brevity, instead of the full Cluedo game\footnote{The reader interested in reviewing the rules of the game can refer to the official 
\href{https://instructions.hasbro.com/en-gb/instruction/clue-the-classic-mystery-game}{instructions} by Hasbro.} we model a simplified version.
Let us consider a game set with 3 players, a dealer, and 8 cards, numbered from 1 to 8.
At the beginning of the game, the dealer samples secretly and puts aside two cards, shuffles the remaining cards together, making sure none of the cards are seen by any of the players, and then deals two cards per player. 
Then, the game starts and proceeds by sequential turns. On her turn, each player makes publicly a suggestion of the form: \textit{I suggest that the two secret cards of the dealer are $i$ and $j$}. There are no constraints about the specific choice of $i$ and $j$.
Then, if the player on the right of the one making the suggestion has at least one of the cards mentioned, she must show one of these cards secretly to her.
Then, the inquiry passes to the player on the left, with the same rule.
At the end of her turn, the player wins the game if she has learned and can correctly declare what the dealer's cards are. Otherwise, the game proceeds with the following turns until one of the players wins.

Formally, we model the game set through the propositions $p_i^j$ and $q_i$, for $0 \le j \le 2$ and $1 \le i \le 8$, where $p_i^j$ means that player $j$ has card $i$ and $q_i$ means that card $i$ is one of the two secret cards of the dealer.
The pool includes one dealer and three players and, initially, is defined as the tuple:
$(\{\langle \mathrm{Mr.~Black}, \mathit{Dealer} \rangle$, 
$\langle 0, \mathit{Player}(0) \rangle$,
$\langle 1, \mathit{Player}(1) \rangle$,
$\langle 2, \mathit{Player}(2) \rangle\},R,X)$.
The three players have the same behavioral pattern, given by the process term $\mathit{Player}$, which is fed with a parameter representing the player identity. 
Set $X$ is empty (the cards have yet to be shuffled by the dealer Mr.\ Black). The accessibility relation of the dealer, 
$R_\mathit{Mr.Black}$, contains only the reflexive pairs, i.e.,
each possible world is a singleton. In fact, by assumption, the dealer is like an oracle and can distinguish any possible scenario. As we will see, $R_\mathit{Mr.Black}$ is immutable.
The accessibility relation for each player $j$, denoted $R_j$, is such that two possible worlds are related if and only if they coincide for the values of the propositions $p^j_i$, $1 \le i \le 8$. The intuition is that, at least, a player is able to distinguish two possible worlds differing in the values of the cards she receives.
All such accessibility relations are equivalence relations but are not immutable, as the knowledge of the players will change as the game proceeds.


Initially, the dealer shuffles the cards and chooses nondeterministically the two secret cards and the assignments for the players (see actions \textit{set}):
\[\begin{array}{l}
\mathit{Dealer} :=\sum_{k_1,k_2} {\mathit{set}}(q_{k_1},1) . {\mathit{set}}(q_{k_2},1) .
\mathit{Deal}(k_1,k_2) \\
\mathit{Deal}(x,y) :=
\sum_{i_1,i_2\not\in\{x,y\}} {\mathit{set}}(p_{i_1}^0,1) .
{\mathit{set}}(p_{i_2}^0,1) . \overline{\mathit{deal}}(0, p_{i_1}^0 \land p_{i_2}^0) . ( \\
\hspace{21mm} \sum_{i_3,i_4\not\in\{i_1,i_2,x,y\}} {\mathit{set}}(p_{i_3}^1,1) .
{\mathit{set}}(p_{i_4}^1,1) . \overline{\mathit{deal}}(1, p_{i_3}^1 \land p_{i_4}^1) . ( \\
\hspace{21mm} \sum_{i_5,i_6\not\in\{i_1,\ldots,i_4,x,y\}} {\mathit{set}}(p_{i_5}^2,1) .
{\mathit{set}}(p_{i_6}^2,1) . \overline{\mathit{deal}}(2, p_{i_5}^2 \land p_{i_6}^2) . \mathit{Play(0)} ) )
\end{array}\]
Whenever clear from the context, the bounds of a summation are not specified (in general, $\sum_{i,j}$ expresses a choice over all the possible unordered pairs of different values $(i,j)$, each one ranging from 1 to 8). Process term $\mathit{Dealer}$ models the random sampling of the two secret cards, and then the invocation of process term $\mathit{Deal}(k_1,k_2)$ describes the following behavior of the dealer whenever $k_1$ and $k_2$ have been chosen. The sampling for each player is modeled analogously through a pair of subsequent actions $\mathit{set}$. The output action $\mathit{deal}$ is used to communicate the assignments to the players.  
Then, the dealer coordinates the game rounds:
\[\begin{array}{l}
\mathit{Play(x)} := \overline{\mathit{start\_turn}}(x,\top) .
(\mathit{end\_turn}(\_,\_) . \mathit{Play((x+1)\mathrm{mod}\,3}) + \mathit{win}(\_,\_) . \underline0) 
\end{array}\]
by assigning each turn (through the output action $\mathit{start\_turn}$) to a different player, sequentially. Note that the output is sent to player $x$ to inform that her turn is starting, without the need to communicate any other information (this justifies the choice of the truth constant $\top$). Then, the dealer waits for a response: either the player turn terminates (input action $\mathit{end\_turn}$) or the player wins the game by 
learning the secret pair during her turn (input action \textit{win}).
For the sake of convenience, whenever unnecessary, the arguments of an input action are left unspecified  (symbol $\_$). 

After receiving the cards through the input action $\mathit{deal}$, each player is available to start her turn (input action $\mathit{start\_turn}$) or to manage inputs from the other players. The process term $\mathit{Player}(x)$ is defined as follows:
\[\begin{array}{l}
\mathit{Player}(x) := \mathit{deal}(y,\_) . \\ \hspace{4mm}
(\mathit{start\_turn}(\_,\_) .
\sum_{i_1,i_2}\overline{\mathit{ask}_{i_1,i_2}}((x+1)\mathrm{mod}\,3,\top) . \mathit{show}(\_,\_) . \\
\hspace{39mm} \overline{\mathit{ask}_{i_1,i_2}}((x+2)\mathrm{mod}\,3,\top) . \mathit{show}(\_,\_) . \\
\hspace{28mm}
(
\overline{\mathit{end\_turn}}(y,\neg \phi_x) . \mathit{Player}(x) +
\overline{\mathit{win}}(y,\phi_x) . \underline0
) 
)
\\ \hspace{4mm}
+ \sum_{i_1,i_2}\mathit{ask}_{i_1,i_2}(z,\_) . \\
\hspace{12mm}(\overline{\mathit{show}}(3-x-z,p_{i_1}^x \vee p_{i_2}^x) . (
\overline{\mathit{show}}(z,p_{i_1}^x) . \mathit{Player}(x) \,+ \\
\hspace{155pt}
\overline{\mathit{show}}(z,p_{i_2}^x) . \mathit{Player}(x)) \, + \\
\hspace{12mm}
\overline{\mathit{show}}(3-x-z, \neg p_{i_1}^x \land \neg p_{i_2}^x) . \overline{\mathit{show}}(z, \neg p_{i_1}^x \land \neg p_{i_2}^x) . \mathit{Player}(x)
) \\ \hspace{4mm}
+ \mathit{show}(\_,\_) . \mathit{Player}(x)
)
\end{array}\]
When initiating a new turn, the player chooses nondeterministically two cards to be asked to each other player (output action \textit{ask}) and then waits for the related answer (input action \textit{show}).
At the end of the turn, either the player learns the secret and wins the game (output action \textit{win}) or passes the hand (output action $\mathit{end\_turn}$). The winning condition for player $x$ determining which output is executed is given by the knowledge of the formula $\phi_x = \bigvee_{(k,k')} K_x (q_k \land q_{k'} )$, i.e., the player knows the secret pair.
Then, players respond to incoming requests through the input action \textit{ask}. If player $x$ receives from player $z$ a request about cards $i_1$ and $i_2$, then we distinguish two cases. Firstly, $x$ may have at least one of the two cards ($p_{i_1}^x \vee p_{i_2}^x$). In this case, $x$ reveals one of the possessed cards to $z$, by choosing the card nondeterministically if necessary. Indirectly, even the third, silent player (identified by $3-x-z$) learns something, i.e., the fact that $x$ has one of the two cards. We model this indirect transfer of knowledge through an explicit output directed to player $3-x-z$. Secondly, $x$ may have none of the two cards ($\neg p_{i_1}^x \land \neg p_{i_2}^x$). In this case, the information is shared with both the other players. Finally, due to the outputs directed to player $3-x-z$, players must also be available to learn some information during the turns of the other players (through the input action $\mathit{show}$).

It is worth noting that the management of the knowledge base of the players is left to the semantics of the underlying Kripke model. At the level of the specification, only the initial setting and the communications are modeled explicitly. This is particularly significant from the viewpoint of usability, as an analogous model based on, e.g., classical Kripke structures, would be much more challenging. To appreciate this aspect, the same use case has been modeled in the software tool NuSMV \cite{10.5555/647771.734431}, the specifications of which result in finite state machines that turn out to be Kripke structures.\footnote{The specification can be found on \href{https://github.com/aldinia/cluedo}{github}.}
Since there are 2520 ways of dealing the 8 cards to the three players and the dealer -- the computing formula is $\binom{8}{2} \cdot \binom{6}{2} \cdot \binom{4}{2}$ -- the NuSMV specification refers to one of these, chosen deterministically through external parameters that initialize the system configuration.
Moreover, the NuSMV specification describes only a very simplistic version of the players' knowledge, in which each player does not deduce any information when observing the interactions between the other two players. 
In fact, the additional information needed to model the full deduction capabilities of the players should be represented explicitly by the designer and would make the model much more complicated and error-prone.
By the way, despite these simplifications, the NuSMV specification is made out of about $200$ code lines and $58$ variables.\footnote{The underlying Kripke structure has about $2^{20}$ states.}

To show an example of properties that can be model checked, we consider the derived \textit{eventually} modality $F$, such that $M,s \models F \phi$ if and only if $M,s \models \phi$ or $\exists \pi \ldotp M,s \models \langle \pi \rangle F \phi$, and the
derived \textit{globally} modality $G$, such that
$M,s \models G \phi$ if and only if $M,s \models \phi$ and $\exists \pi \ldotp M,s \models \langle \pi \rangle G \phi$.
Then, the reachability property $F (\bigvee_x \phi_x)$ is satisfied, i.e., the \textit{winning} state is reachable by some players. However, even the unreachability property $G (\bigwedge_x \neg \phi_x)$ holds. The reason is that
the nondeterministic strategy followed by the players when choosing their suggestion does not guarantee that the game can always be won.


\section{Related work and conclusions}
\label{sect:rw}

A few approaches investigate the combination of LTS-based semantics and epistemic notions, e.g., in the setting of epistemic $\mu$-calculus~\cite{10.1007/978-3-540-75560-9_18} and of concurrent constraint programming paradigms~\cite{10.1007/978-3-642-32940-1_23,GUZMAN2017107}. The framework proposed in~\cite{10.5555/2340820.2340835} is the closest to our approach in principle, as it integrates LTSs with accessibility relations stating the indistinguishability between states. However, agents observe (do not control) the path of performed actions and, based on this knowledge, deduce what the actual state is. Hence, the semantics of the formulas of the underlying logic is given in terms of paths. Notably, such a logic, similarly to the KT logic, is equipped with both temporal and epistemic modalities.

An important strand of research concerns dynamic extensions of Kripke models and epistemic logic, where the dynamic dimension is related to the execution of actions over time; see, e.g., \cite{BaltagMoss2004,VANOTTERLOO200577,10.1145/2000378.2000403,10.1145/1324249.1324262,Dim2007,GSL2012,Baltag2016,a81d0938-6dc3-3b28-a7ca-55199276cfea,Bolander_2017} and the references therein. 
However, all these approaches differ in the way in which we encode the dynamics of epistemic models within the LTS-style semantics. The main advantage of our encoding is that the obtained semantics facilitates the definition of a high-level process-algebraic language for the description of multi-agent systems and knowledge-based interactions.

In the field of concurrency theory, some of the ideas presented in this paper can be found in the study of temporal logics encompassing features from HML and modal $\mu$-calculus~\cite{10.1007/3-540-53479-2_17}. As an example, a variant of the temporal logic CTL is defined in~\cite{TERBEEK2011119} to check properties over expressive models called $\mathrm{L}^2\mathrm{TS}$.
In these models, the idea is to combine transition labels expressing the action-based dynamic behavior of a system with state-based labels expressing the knowledge possessed in each state of the system. With respect to our proposal, no epistemic representation of derivable knowledge is given, so the study of the observational power of the agents is limited to the verification of state-based propositional logic formulas and on the model checking of temporal formulas. 




Summarizing, by following suggestions deriving from works on dynamic and temporal epistemic logics~\cite{ParRam2003}, we embedded a structure of pointed Kripke models into a labeled transition system, the actions of which act as model-transforming operations from the viewpoint of the Kripke models.
These transitions naturally model the behavior of the system and the passage of (discrete) time, while the Kripke models linked to the states visited during the temporal evolution of the system represent the way in which the knowledge of every agent evolves over time.
The process algebraic language that we introduced emphasizes these effects and allows for a compact and elegant description of multi-agent systems, where the details of the knowledge evolution are left to the underlying epistemic model.

Starting from this point, several extensions can be envisioned. For instance, the semantics of our communication mechanisms assumes that only known truth can be transferred. Hence, we do not currently manage (possibly false) beliefs and the communication of information 
that is inconsistent with an agent’s knowledge or belief. This would require the introduction of the belief modality and the treatment of contradictions resulting from the communication between agents. Moreover, this would also open to extensions in which it is possible to model the behavior of malicious agents sharing false information and, therefore, a theory of fake news \cite{8560741,DBLP:journals/logcom/Aldini22}. Along the same lines, further modalities could be added to the epistemic component of our model.

Dealing with inconsistencies is a problem to face even in the present model, without bringing up the notion of belief. In particular, an unsuccessful formula is a formula that might become false as soon as it is communicated, like, e.g., in the case of $p \land \neg K_j p$ whenever agent $i$ communicates it to agent $j$ \cite{DimKooi2006}. Several studies investigate the syntactic form of potential unsuccessful formulas, in particular in the setting of public announcements for multi-agent systems \cite{DBLP:journals/corr/abs-1209-0935}. 
Obviously, even in our framework such forms can be recognized and, in particular, are limited to those cases in which a formula of the form $\neg K_j \psi$ is involved in a communication to agent $j$. This is because the satisfaction of such a formula before the communication could be contradicted by sharing its knowledge with the agent suffering from such a kind of ignorance.

Finally, in order to expand the theoretical development of our framework, we also plan to define: $(i)$ an axiomatization for the KT logic, $(ii)$ quantitative extensions of the KLTS model, by adding continuous time and probabilistic choices, and $(iii)$ additional ingredients in the process-algebraic language, by including internal actions guarded by knowledge-based conditions, \textit{if-then-else} constructs that are based on knowledge conditions, and broadcast communication in the style of~\cite{Ald18}. 
%
%
%
 \bibliographystyle{splncs04}
 \bibliography{mybibliography}

\appendix
\section{Proofs of results}

\subsection{Theorem~\ref{main_thm}}

The proof of Theorem~\ref{main_thm} is an adaptation of standard approaches \cite{blackburn2002modal,HML}. Consider the KLTSs 
$M_1 = (S_1,T_1,\{r_{1i} \mid i \in \mathcal{A}\},v_1)$ pointed at $s_{01}$ and $M_2 = (S_2,T_2,\{r_{2i} \mid i \in \mathcal{A}\},v_2)$ pointed at $s_{02}$. 

Case ($\Rightarrow$): $s_{01} \sim s_{02}$ implies $s_{01} \equiv s_{02}$. \\
Let us assume that there exists a bisimulation $B$ including the pair $(s_{01},s_{02})$. We will show the result through an induction on the structure of the formulas.

The base case refers to the atomic formulas and trivially holds by definition of bisimulation since $v_1(s_{01}) = v_2(s_{02})$. Hence, by applying the induction hypothesis and the semantics of the propositional logic operators, no propositional formula can distinguish bisimilar states. 

In the case of $\langle \pi \rangle \phi$, suppose that $(M_1,s_{01}) \models \langle \pi \rangle \phi$ because $\exists s_1 \in S_1$ such that $(s_{01},\pi,s_1) \in T_1$ and $M_1, s_1 \models \phi$. Hence, by hypothesis, 
$\exists s_2 \in S_2$ such that $(s_{02},\pi,s_2) \in T_2$ and $(s_1,s_2) \in B$. By applying the induction hypothesis, it holds that $M_2,s_2 \models \phi$, so that no temporal formula $\langle \pi \rangle \phi$ can distinguish $s_{01}$ from $s_{02}$.

In the case of the epistemic formula $K_i \psi$, suppose that $(M_1,s_{01}) \models K_i \psi$ because $M_{1{s_{01}}}, v_1(s_{01}) \models_{\mathrm{K}} K_i \psi$. 
By hypothesis, $(2^{At},r_{1i}(s_{01}),id)$ pointed at $v_1(s_{01})$ and
$(2^{At},r_{2i}(s_{02}),id)$ pointed at $v_2(s_{02})$ are related by a binary equivalence relation $\mathcal{B}$ including $(v_1(s_{01}), v_2(s_{02}))$ as given in Definition~\ref{def-bisimulation}. We now show that such models satisfy the same epistemic formulas and, therefore, $M_{2s_{02}}, v_2(s_{02}) \models_{\mathrm{K}} K_i \psi$, from which we derive $(M_2,s_{02}) \models K_i \psi$.
\begin{itemize}
\item Base: Since $v_1(s_{01}) = v_2(s_{02})$, $M_{1s_{01}}, v_1(s_{01}) \models_{\mathrm{K}} p$ iff $M_{2s_{02}}, v_2(s_{02}) \models_{\mathrm{K}} p$.
\item Induction: Trivially, no propositional formula can distinguish $v_1(s_{01})$ from $v_2(s_{02})$. So, let us consider $M_{1s_{01}}, v_1(s_{01}) \models_{\mathrm{K}} K_i \psi$. Now, let $X$ be such that $(v_1(s_{01}),X) \in r_{1i}(s_{01})$, hence it must be that $M_{1s_{01}}, X \models_{\mathrm{K}} \psi$. By hypothesis,
there exists $Y$ such that $(v_2(s_{02}),Y) \in r_{2i}(s_{02})$ with $(X,Y) \in \mathcal{B}$ and, by induction hypothesis, $M_{2s_{02}}, Y \models_{\mathrm{K}} \psi$. Since this holds for every world accessible from $v_1(s_{01})$, it holds that $M_{2s_{02}}, v_2(s_{02}) \models_{\mathrm{K}} K_i \psi$, as expected.
\end{itemize}


Therefore, having a bisimulation relating two states is sufficient for the two states to verify the same KT formulas.

Case ($\Leftarrow$): $s_{01} \equiv s_{02}$ implies $s_{01} \sim s_{02}$. \\
We will show (by contradiction) that $\equiv$ is a bisimulation itself. 

First, the condition $v_1(s_{01}) = v_2(s_{02})$ trivially holds because its violation would mean that there exists a propositional formula distinguishing the two states. 

Second, take an arbitrary action $\pi$ and assume that there exists $s_1$ such that $(s_{01},\pi,s_1) \in T_1$, but there does not exist $s_2$ such that $(s_{02},\pi,s_2) \in T_2$ with $s_1 \equiv s_2$. Let $S'$
be the finite set of states accessible from $s_{02}$ through a $\pi$-labeled transition. $S'$ is non-empty otherwise $\langle \pi \rangle \top$ would distinguish $s_{01}$ from $s_{02}$. By assumption, for each $t_j \in S'$, $1 \le j \le |S'|$, there exists $\phi_j$ such that $M_1, s_1 \models \phi_j$ and $M_2, t_j \not\models \phi_j$.
Hence, it holds that $M_1, s_{01} \models \langle \pi \rangle \bigwedge_j \phi_j$ (since there is $s_1$ such that $(s_{01},\pi,s_1) \in T_1$ and satisfying $\bigwedge_j \phi_j$), and $M_2, s_{02} \not\models \langle \pi \rangle \bigwedge_j \phi_j$, thus contradicting the hypothesis. The same kind of reasoning applies to the symmetric case.

Third, take an arbitrary $r_{1i}(s_{01})$ and assume that $(2^{At},r_{1i}(s_{01}),id)$ pointed at $v_1(s_{01})$ and $(2^{At},r_{2i}(s_{02}),id)$ pointed at $v_2(s_{02})$ are not related by a binary equivalence relation $\mathcal{B}$ including $(v_1(s_{01}), v_2(s_{02}))$ as given in Definition~\ref{def-bisimulation}. As a first consideration, note that
$v_1(s_{01}) = v_2(s_{02})$, hence we assume that there exists $X$ such that
$(v_1(s_{01}), X) \in r_{1i}(s_{01})$, but there does not exist $Y$ such that $(v_2(s_{02}), Y) \in r_{2i}(s_{02})$ with $(X,Y)$ in $\mathcal{B}$. Let $S'$ be the finite set of worlds accessible from $v_2(s_{02})$ through $r_{2i}(s_{02})$. $S'$ is non-empty, 
otherwise $K_i \neg \top$ would distinguish $v_1(s_{01})$ from $v_2(s_{02})$.
By assumption, for each $Y_j \in S'$, $1 \le j \le |S'|$, there exists $\phi_j$ such that $M_{1s_{01}}, X \models_{\mathrm{K}} \phi_j$ and $M_{2s_{02}}, Y_j \not\models_{\mathrm{K}} \phi_j$.
Hence, it holds that $M_{1s_{01}}, v_1(s_{01}) \models_{\mathrm{K}} \neg K_i \neg \bigwedge_j \phi_j$ (since there is $X$, accessible from $v_1(s_{01})$, satisfying $\bigwedge_j \phi_j$) and 
$M_{2s_{02}}, v_2(s_{02}) \not\models_{\mathrm{K}} \neg K_i \neg \bigwedge_j \phi_j$
(since by assumption and the semantics of the epistemic operator, $M_{2s_{02}}, v_2(s_{02}) \models_{\mathrm{K}} K_i \neg \bigwedge_j \phi_j$), thus contradicting the hypothesis. The same kind of reasoning applies to the symmetric case.

Therefore, the modal equivalence $\equiv$ satisfies the three conditions of Definition \ref{def-bisimulation} and, hence, is a bisimulation, thus completing the proof.

\subsection{Theorem~\ref{rel-thm}}

The proof proceeds by showing that given any state $s \in S$ such that, for each $i \in \mathcal{A}$, $r_i(s)$ is a \texttt{P}-relation, then every state $s'$ reachable from $s$ through a transition $(s,\_,s') \in T$ satisfies the same \texttt{P}-condition. Then, the result immediately follows from the hypothesis that the initial state satisfies the \texttt{P}-condition.

By following the rules of Table~\ref{tab:sem_pool}, we first observe that the rule $(\textit{pool})$ does not change the relation set $\{R_i \mid i \in \mathcal{A} \}$. Hence, the non-trivial cases are the rules (\textit{set}) and (\textit{com}), where the latter is a sub-case of the former as far as the updates to the relation set are concerned. Hence, we now show the effect of an action of the form \textit{set}($p,\_$).

Let us assume that the relation set $\cup_i R_i \cup R_j$ in $s$ is turned into the relation set $\cup_i R'_i \cup R'_j$ in $s'$.

By definition, $R'_j$ derives from $R_j$ by deleting those pairs $(Y,Y')$ such that $Y$ and $Y'$ differ for the evaluation of $p$. Hence, it is easy to see that reflexivity and symmetry are preserved. Let us consider transitivity, by assuming that there exist $Y,Y',Y''$ such that $(Y,Y'),(Y',Y''),(Y,Y'') \in R_j$ but only $(Y,Y'),(Y',Y'') \in R'_j$. Let $(Y,Y'') \not\in R'_j$ since $p \in Y$  and $p \not\in Y''$ (the opposite case is orthogonal). Then, $p \in Y'$ contradicts the assumption $(Y',Y'') \in R'_j$, while $p \not\in Y'$ contradicts the assumption $(Y,Y') \in R'_j$. Hence, it cannot be that $(Y,Y'),(Y',Y'') \in R'_j$ and $(Y,Y'') \not\in R'_j$. 

By definition, for any $i$ we have that $R'_i$ derives from $R_i$ by adding those pairs $(Y,Y')$ such that $Y$ and $Y'$ differ only for the evaluation of $p$. Hence, it is easy to see that reflexivity and symmetry are preserved. Let us consider transitivity, by assuming that there exist $Y,Y',Y''$ such that $(Y,Y') \in R_i$, $(Y',Y'') \not\in R_i$ and $(Y,Y'), (Y',Y'') \in R'_i$. Hence, by the \textit{closure} operation, we have that $(Y,Y'') \in R'_i$ too.

\end{document}